\documentclass{article}
\usepackage{kr}

\usepackage{times}
\usepackage{soul}
\usepackage{url}
\usepackage[hidelinks]{hyperref}
\usepackage[utf8]{inputenc}
\usepackage[small]{caption}
\usepackage{graphicx}
\usepackage{amsmath}
\usepackage{amsthm}
\usepackage{booktabs}

\usepackage{tabularx}
\usepackage{algorithm}
\usepackage{amssymb}
\usepackage{enumitem}
\usepackage{algpseudocode}
\usepackage{natbib}
\usepackage{multirow}

\title{Probabilistic Active Goal Recognition}

 \author{%
 Chenyuan Zhang\and
 Cristian Rojas Cardenas\and
 Hamid Rezatofighi\and
 Mor Vered \and
 Buser Say\\
 \affiliations
 Monash University
 \emails
 \{chenyuan.zhang, cristian.rojascardenas, hamid.rezatofighi, mor.vered, buser.say\}@monash.edu
 }

\begin{document}

\maketitle

\begin{abstract}
  In multi-agent environments, effective interaction hinges on understanding the beliefs and intentions of other agents. While prior work on goal recognition has largely treated the observer as a passive reasoner, Active Goal Recognition (AGR) focuses on strategically gathering information to reduce uncertainty. We adopt a probabilistic framework for AGR and propose an integrated solution that combines a joint belief update mechanism with a Monte Carlo Tree Search (MCTS) algorithm, allowing the observer to plan efficiently and infer the actor's hidden goal without requiring domain-specific knowledge. Through comprehensive empirical evaluation in a grid-based domain, we show that our joint belief update significantly outperforms passive goal recognition, and that our domain-independent MCTS performs comparably to our strong domain-specific greedy baseline. These results establish our solution as a practical and robust framework for goal inference, advancing the field toward more interactive and adaptive multi-agent systems.
\end{abstract}

\section{Introduction}

Imagine a robot assistant working alongside a human in a manufacturing environment. The human may be assembling one of several different products, each requiring a distinct assembly sequence. To provide effective assistance, the robot must infer the human’s intended step as early as possible. This scenario highlights a broader challenge in human–robot collaboration: successful interaction often depends on accurately understanding the intentions and beliefs of other agents~\citep{demiris2007prediction,dann2023multi}.

While crucial for effective multi-agent interaction, the problem of recognizing other agents’ goals and modeling their beliefs has received relatively limited attention in much of the multi-agent systems literature. Many studies in this area rely on model-free or learning-based approaches that do not explicitly reason about other agents~\citep{zhang2021multi,canese2021multi}, or defer such reasoning to large language models~\citep{li2023theory,shi2025muma}. While these methods can perform well in reactive or end-to-end tasks, they often lack interpretability and struggle in situations that require anticipating others’ intentions or long-term behavior. This limitation highlights the need for multi-agent frameworks that explicitly incorporate goal and belief modeling into decision-making—particularly in domains where understanding other agents is key to effective interaction.

In contrast, the goal recognition community typically employs symbolic models and planning-based approaches to reason about agent behavior~\citep{masters2019cost,vered2018towards,ramirez2010probabilistic}. While these studies differentiate between a range of observed agents, be it unaware of being observed or even deceptive, they uniformly assume that the observer is a passive entity that cannot affect the environment and focuses strictly on inferring goals from a sequence of observations. However, in many real-world scenarios the observer is not merely a passive reasoner but an agent capable of taking actions to shape its own information state~\citep{fitzpatrick2021behaviour}. This aligns with the field of active information gathering, which studies how to select actions that reduce uncertainty~\citep{shah2014collaborative,veiga_reactive_2023}. 

By unifying passive goal recognition with active information gathering, \cite{amato2019active} introduced the problem of \emph{Active Goal Recognition} (AGR). In their formulation, the observer is tasked with both recognizing the actor's goal and completing its own planning objective, requiring a balance between task execution and information gathering. Although they model the problem as a Partially Observable Markov Decision Process (POMDP), they manually design the reward function instead of deriving it from the formulation.
Around the same time, \cite{shvo2020active} also proposed an AGR formulation using the STRIPS-like language. Their approach leverages a landmark-based planning algorithm to actively collect observations for goal recognition. This method closely aligns with traditional goal recognition as planning approaches and does not incorporate a probabilistic formulation.

In this work, we focus on a setting where the observer’s sole objective is to recognize the actor’s hidden goal, without pursuing any independent task. To model this, we introduce a Probabilistic Active Goal Recognition (PAGR) framework based on a POMDP formulation. This framework leverages structured knowledge representation and belief-based rewards, enabling the observer to reason and act under uncertainty. Our formulation provides a unified probabilistic and decision-theoretic perspective to address a central question: how should an observer act in the environment to actively uncover the actor’s goal?

\section{Related Work}

In this section, we provide an overview of the previous works that have modeled and solved related problems.

\subsection{Goal Recognition}
Two prevalent approaches of goal recognition are either using plan libraries or plan recognition as planning (PRP)~\citep{meneguzzi_survey_2021}. Algorithms based on plan libraries, also known as plan recognition as parsing, present plans as a hierarchy of simpler actions. The main task becomes aligning the observed actions with these structured plans. Hierarchical Task Networks (HTNs) and grammars are typical methods for representing knowledge in plan libraries~\citep{stuart2016artificial}. HTN outlines tasks using a set of subtasks and their constraints, either separately or in relation to each other. Meanwhile, grammars describe the structure of plans through a set of production rules. These algorithms are useful in domains where the set of possible plans is known in advance, such as in video game AI and robotics~\citep{van-horenbeke_activity_2021}.

On the other hand, PRP approaches use standard planning algorithms to create potential plan hypotheses for the observed agent ~\citep{ramirez2010probabilistic, vered2016online, vered2017heuristic, zhang2023goal, kaminka2018plan, masters2019cost}. These planning algorithms are typically formulated using planning languages like STRIPS or PDDL, enabling them to outline the state of the environment and the impacts of applicable actions. In these approaches planners are used to calculate potential plans, as needed,  that could achieve specified goals from varied initial states. The plan recognition system then assigns weights to these candidate plans by matching them against incoming observations, and the most likely plan or goal is chosen based on these weights. 

In their survey, \cite{ijcai2021p615} identified the implicit and explicit assumption made by goal recognition researchers. A common limitation that emerges across all prior works is the implicit assumption that the observer is a static agent and unable to change its state to improve its goal recognition capability.

\subsection{Active Goal Recognition}
Active Goal Recognition, unlike standard goal recognition, involves an observer that can influence the observation process through its own actions. The observations of the observing agent depend on the actions it takes within its own domain, which may differ from the domain of the target. This allows the observer to strategically gather information so as to improve the accuracy and efficiency of goal inference~\citep{shvo2020active}.

\cite{shvo2020active} was the first to formalize the AGR problem and proposed a landmark-based approach for solving it. Their method performs hypothesis elimination within a partially observable planning framework grounded in STRIPS. Around the same time, \cite{amato2019active} introduced a more general framework based on POMDPs, enabling the handling of stochastic actions and transitions. Their solution relies on a linear approximation using the SARSOP solver and requires a manually designed reward structure, which depends heavily on domain knowledge. In a related line of work, \cite{ijcai2021p559} studied Goal Recognition Design (GRD) in an active setting, where the observer can interact with and modify the environment to induce the actor to reveal their true goal earlier. However, their formulation assumes full observability, and the observer’s actions aim at breaking path symmetries rather than gathering information.

In this work, we propose Probabilistic Active Goal Recognition (PAGR) to address the limitations of prior approaches. Our formulation considers a partially observable setting where the observer must act to collect information and reduce uncertainty over the actor’s goal, within a unified probabilistic and decision-theoretic framework.

\subsection{Active Information Gathering and Online POMDP Solvers}

To effectively perform AGR in partially observable environments, the observer must continuously update its beliefs and select informative actions in real time. This aligns with the broader literature on active information gathering and online POMDP solving. In this section, we review key approaches and solvers relevant to this approach.

Traditional information gathering has mostly been viewed as a means to an end in planning problems under uncertainty. In contrast, \textit{active} information gathering refers to scenarios in which acquiring information about the environment or other agents is an integral part of the system's objective~\citep{bajcsy_active_1988}. While standard reactive sensing relies on decisions driven by observed data, active information gathering, also known as active sensing, addresses this challenge by developing strategies that incorporate reasoning, decision-making, and control to maximize the value of the information collected~\citep{veiga_reactive_2023}.

These problems are usually modeled as POMDPs. Under the extension of uncertainty, the model is modified with a belief-based reward rather than a state-based one, resulting in what is known as a $p$-POMDP. However, $p$-POMDPs are computationally more expensive than normal POMDPs due to the introduction of the belief space~\citep{NIPS2010_68053af2}. 

Due to the intractability of solving POMDPs offline in complex domains, online POMDP solvers have been developed over the past two decades to enable scalable planning. A prominent example is POMCP~\citep{silver2010monte}, which applies Monte Carlo Tree Search (MCTS) to sample and evaluate future trajectories from the current state without modeling the belief space. Building on this approach, many variants have been proposed, differing in components such as backup strategies and sampling mechanisms~\citep{sunberg2018online, thomas_vincent_monte_2020}. To further improve scalability, especially in continuous state or action spaces, \cite{sunberg2018online} introduced an extension of POMCP that supports planning in continuous domains, greatly expanding the applicability of online solvers to $p$-POMDPs. These advances provide a practical foundation for active decision-making in our setting.

\section{The Probabilistic Active Goal Recognition Problem}\label{problem_statement}



\begin{figure}
    \centering
    \includegraphics[width=0.5\linewidth]{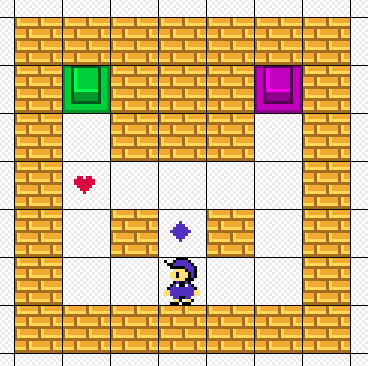}
    \caption{Illustrated example for Active Goal Recognition.}
    \label{fig:ep}
\end{figure}

To motivate the problem, consider a simplified scenario where a single actor navigates an environment to reach one of two possible goals. An observer, aiming to identify the actor’s true goal as efficiently as possible, is allowed to place a monitor in just one location. As shown in Figure~\ref{fig:ep}, the two potential goals are marked by green and purple boxes. The observer knows the actor’s starting position and that the true goal is one of the two possibilities, but receives no information about the actor’s movements once the task begins except through the monitor, mimicking the limited observability available in real-world settings. The monitor provides a single observation by triggering if the actor passes through the chosen location. The challenge for the observer is to strategically place the monitor so as to maximize the chance of correctly inferring the actor’s goal. 

Let us examine two potential monitor placements, indicated by the red heart and blue diamond. Although the actor is very likely to pass through the blue diamond cell, this observation may offer little value: it lies along the optimal path to both potential goals, and thus provides minimal disambiguation. In contrast, placing the monitor on the red heart cell will yield a more informative observation. If the actor passes through it, this strongly indicates the green goal is the intended destination. Conversely, if the actor does not trigger the monitor at the red cell, this absence of evidence counts as negative evidence against the green goal, increasing the likelihood that the purple goal is the true target. This example shows the importance of selecting observer actions that maximize informativeness for goal disambiguation.

\subsection{Formal Definition}

We consider an environment shared by two agents: an \emph{actor} and an \emph{observer}. The actor is engaged in solving a planning problem to achieve a hidden goal, while the observer aims to infer this goal, as early as possible, while potentially interacting with the environment. We assume keyhole GR whereby the actor is unaware of, and unaffected by, being observed~\citep{ijcai2021p615}; that is, the actor’s policy and state transitions are independent of the observer’s state or actions. This assumption simplifies the formulation and is consistent with prior work in passive goal recognition. Extending the framework to model interactive actors is an important direction for future research, see Section~\ref{sec:discussion}.


We build on the Active Goal Recognition (AGR) formulation introduced by \cite{amato2019active}, which allows for various types of observer actions, while the deterministic observation model is limited to specific types of observations that directly relate to the actor. In this work, we adapt and generalize their formulation to create a more concise and broadly applicable formulation. In particular, we redefine the observation function to depend jointly on the states of both the actor and the observer. This modification enables a more flexible and expressive model of perceptual uncertainty, accommodating a wider range of observation scenarios. We note that our formulation falls within the factored DEC-POMDP framework~\citep{oliehoek2016concise}, which captures structured multi-agent decision making under partial observability.

In the environment, the actor solves the following planning problem $\mathcal{P}_{actor} = \langle s_0,\,\mathcal{E}_{actor},\,g^*\rangle$, where the environment dynamics $\mathcal{E}_{actor} = \langle \mathcal{S},\,\mathcal{A}^A,\,f^A\rangle$ are determined by the actor’s state space $\mathcal{S}$, the initial state $s_0 \in \mathcal{S}$, the actor's action space $\mathcal{A}^A$, the transition function $f^A(s,a^A,s') =  P(s'\mid s,a^A)$ and the goal state $g^* \in \mathcal{S}$.
Given the actor planning problem $\mathcal{P}_{actor}$, the observer performs AGR $\mathcal{P}_{PAGR} = \langle \mathcal{E}_{actor}, \mathcal{U}, u_0, \mathcal{A}^O,  f^O, \mathcal{O},f_{\text{obs}}, \mathcal{G} \rangle$ where:
\begin{itemize}
\item $\mathcal{U}$ is the state space of the observer.
    \item $u_0 \in \mathcal{U}$ is the initial state of the observer.
    \item $\mathcal{A}^O$ is the observer's action space.
    \item $f^O(u, a^O, u'): \mathcal{U} \times \mathcal{A}^O \times \mathcal{U} \rightarrow [0, 1]$ be the observer's transition probability function, representing the probability of transitioning from state $u$ to $u'$ given an observer action $a^O$.
    \item $\mathcal{O}$ is the observation space.
    \item $f_{\text{obs}}(u, s, o): \mathcal{U} \times \mathcal{S} \times \mathcal{O} \rightarrow [0, 1] $ denote the observation function, specifying the probability of observing \( o\in \mathcal{O} \) given observer state \( u \) and actor state \( s \). For simplicity, we assume that it depends only on the current states.
    \item $\mathcal{G} \subseteq \mathcal{S}$ is the candidate goal set such that $g^* \in \mathcal{G}$.
\end{itemize}

Here, the observation space $\mathcal{O}$ (e.g.\ ${\text{detect},\text{not detect}}$ in the motivated example) captures the fact that the observer never sees $s$ directly but must infer it via these partial, state‑dependent signals from the observation function $f_{\text{obs}}$. Unlike prior AGR formulations that treat sensing/observing as a distinct action class, our framework assumes a more general setting in which every observer action may influence the distribution of its subsequent observations. Note that while the observer has access to the actor's environment dynamics \( \mathcal{E}_{actor} \), it does not know the actor's specific goal \( g^* \) or the initial state \( s_0 \). Meanwhile, it is assumed to know the set of candidate goals \( \mathcal{G} \).


At each time step \( t \), the actor is in state \( s_t \in \mathcal{S} \). The actor’s state evolves stochastically according to the transition dynamic \( P(s_{t+1} \mid s_t, a^A_t) \), where actions \( a^A_t \in \mathcal{A}^A \) are selected based on a (possibly stochastic) policy \( \pi^A(a^A_t \mid s_t, g^*) \) that aims to achieve the goal \( g^* \). This policy arises from solving a planning problem conditioned on the goal. Importantly, the observer does not have access to the actor’s policy \( \pi^A \), and cannot observe the actor’s states or actions directly.

The observer maintains its own observer state \( u_t \in \mathcal{U} \), which evolves according to its own transition dynamic $f^O$. While the observer does not have access to the actor’s internal state or policy, it receives a noisy observation \( o_t \in \mathcal{O} \) that provides partial information about the actor’s state. This process is characterized by the observation function $f_{\text{obs}}$.

For notational simplicity, we will use \( a_t \) to denote the observer’s action at time $t$ when no ambiguity arises. 
In \emph{Probabilistic Active Goal Recognition (PAGR)}, the observer aims to infer the actor’s hidden goal \( g^* \in \mathcal{G} \) through a sequence of interleaved actions and observations, which involves two key components. First, at each time step \( t \), the observer maintains a belief distribution over all candidate goals, denoted \( b_t(g) \) for each \( g \in \mathcal{G} \). Second, it must execute a behavior policy to gather informative observations that support belief update.

Quantifying the effectiveness of the observer’s algorithm is non-trivial. In many applications, it is important not only to identify the correct goal but to do so both early and confidently. To formalize this, we adopt the notion of \emph{convergence} (CV), following the formulation introduced by \cite{vered2018towards}, which captures both the timeliness and certainty of goal inference:
\[
\text{CV}(g^*) = 
\begin{cases}
\displaystyle \frac{T - \tau(g^*)}{T}, & \text{if } \exists \, \tau(g^*) \leq T  \\
0, & \text{otherwise}
\end{cases}
\]
where \( \tau(g^*) = \min \left\{ t : \forall t' \in [t, T],\; b_t(g^*) \geq \theta \right\} \), \( T \) is the total task horizon, and \( \theta \) is a predefined threshold. 

The objective of the PAGR problem is to find an observer policy \( \pi^O(a_t \mid u_t, o_{0:t}) \) that selects informative actions, along with a belief update mechanism that maintains \( b_t(g) \) over time, in order to maximize the convergence \( \text{CV}(g^*) \) with respect to the true goal in a stochastic environment.

\section{Inference and Planning for Active Goal Recognition}

To address the PAGR problem, we propose a framework that integrates probabilistic belief update with decision-theoretic planning. In this framework, the observer maintains a joint belief distribution over the actor’s possible goals and states, and selects actions aimed at actively reducing this uncertainty. The belief is updated using a Bayesian inference mechanism informed by the sequence of observations obtained through interaction with the environment. To determine informative actions, we employ a Monte Carlo Tree Search (MCTS) algorithm tailored to the PAGR setting. This section introduces the joint belief update formulation and the MCTS algorithm that guides the observer’s behavior.

\subsection{Joint Belief Update}\label{jbu}

The observer maintains and updates a belief over both the actor’s hidden goal \( g \) and its internal state \( s \) for each time step \( t \). This is represented as a joint belief over the pair \( (s_t, g) \). Note that  the update of this belief depends not only on the history of the observations \( o_{0:t} \), but also on the observer’s own trajectory \( u_{0:t} \), since the observation function $f_{\text{obs}}$ is conditioned on the observer's state. Intuitively, this reflects the fact that the informativeness of each observation depends on the observer’s state, which determines its perspective and sensing capabilities.
To formalize this, we define the joint belief distribution \( j_t \) at time \( t \) as:
\begin{equation}
\label{eq:j_def}
    j_t(s_t, g) = P(s_t, g \mid o_{0:t}, u_{0:t}),
\end{equation}
which expresses the observer's probabilistic estimate of the actor being in state \( s_t \) and pursuing goal \( g \), given the observation and observer state histories. This joint distribution forms a matrix over \( \mathcal{S} \times \mathcal{G} \).

Next, we adopt a sequential update approach, where the belief \( j_t \) is recursively computed from \( j_{t-1} \), using newly acquired information \( (o_t, u_t) \). 
To model the actor’s behavior, we assume it follows a goal-directed policy. Under this assumption, the actor’s state transition can be written as:
\begin{equation}
\label{eq:act_trans}
    P(s_t \mid s_{t-1}, g) = \sum_{a^A} P(s_t \mid s_{t-1}, a^A)\, \hat{\pi}^A(a^A \mid s_{t-1}, g),
\end{equation}
where \( P(s_t \mid s_{t-1}, a^A) \) is the known environment dynamics $f^A$ and \( \hat{\pi}^A(a^A \mid s_{t-1}, g) \) is how observer models actor’s goal-conditioned policy.

Using this, we first perform a prediction step:
\begin{align}
     &P(s_t, g \mid o_{0:t-1}, u_{0:t-1}) \notag \\
        &= \sum_{s_{t-1}} P(s_t, g, s_{t-1} \mid o_{0:t-1}, u_{0:t-1})\label{line1} \\
        &= \sum_{s_{t-1}} P(s_t \mid s_{t-1}, g) P(s_{t-1}, g \mid o_{0:t-1}, u_{0:t-1})\label{line2}\\
        & = \sum_{s_{t-1}} P(s_t \mid s_{t-1}, g) j_{t-1}(s_{t-1}, g)\label{line3}.
\end{align}
Equation~\ref{line1} applies marginalization. Equation~\ref{line2} uses local Markov property of $s_t$, which states that $s_t$ is conditionally independent of past variables given $s_{t-1}$ and $g$; Equation~\ref{line3} is the definition of $j_{t-1}$.

Then, we perform the update step using the current observation \( (o_t, u_t) \) and the observation model \( P(o_t \mid s_t, u_t) \):
\begin{align}
\label{eq:update}
    j_t(s_t, g) &= P(s_t, g \mid o_{0:t}, u_{0:t}) \\
    &= \frac{P(o_t \mid s_t, u_t) \cdot P(s_t, g \mid o_{0:t-1}, u_{0:t-1})}
    {\sum_{s'_t, g'} P(o_t \mid s'_t, u_t) \cdot P(s'_t, g' \mid o_{0:t-1}, u_{0:t-1})}.
\end{align}
This equation applies Bayes’ rule to incorporate the new observation \( o_t \) and observer state \( u_t \). The numerator reflects the product of the observation likelihood \( P(o_t \mid s_t, u_t) \) and the predictive belief \( P(s_t, g \mid o_{0:t-1}, u_{0:t-1}) \) obtained from the previous step. The denominator normalizes the distribution by summing over all possible state-goal pairs \( (s'_t, g') \), ensuring that \( j_t \) is a valid probability distribution. This step also relies on the conditional independence assumption:
\begin{equation}
P(o_t \mid s_t, g, o_{0:t-1}, u_{0:t}) = P(o_t \mid s_t, u_t),
\end{equation}
which states that given the current actor state \( s_t \) and observer state \( u_t \), the observation \( o_t \) is conditionally independent of the goal \( g \), the observation history \( o_{0:t-1} \), and the observer's previous trajectory \( u_{0:t-1} \). This follows from the structure of the observation model and is a consequence of the local Markov property in the underlying graphical model.

Combining both steps yields the full recursive update:
\begin{align}
\label{f_transition}
    &j_t(s_t, g) = \notag \\ & \frac{P(o_t \mid s_t, u_t)\sum_{s_{t-1}} P(s_t \mid s_{t-1}, g) j_{t-1}(s_{t-1}, g)}
    {\sum_{s'_t, g'} P(o_t \mid s'_t, u_t)\sum_{s'_{t-1}} P(s'_t \mid s'_{t-1}, g') j_{t-1}(s'_{t-1}, g')}.
\end{align}

We denote this update compactly as $j_t = h(j_{t-1}, u_t, o_t)$ , indicating that the current joint belief is computed from the previous belief and the latest information.

Once the joint belief distribution \( j_t(s_t, g) \) is obtained, the marginal belief over goals can be computed by summing over the actor's possible states:
\begin{equation}
    b_t(g) = \sum_{s_t} j_t(s_t, g).
\end{equation}

\subsection{Belief-Guided Action Selection}

With the joint belief \( j_t \) computed at each time step, the observer must choose an observer action \( a_t \in \mathcal{A} \) that maximizes its ability to infer the actor’s true goal. A natural reward signal $R(j_t)$ is the marginal belief \( b_t(g^*) \), which quantifies the observer’s confidence in the true goal \( g^* \) under the current belief $j_t$. However, this formulation is not directly usable for action selection, since the observer does not know which goal \( g \) is the true goal.

Instead, the observer can aim to maximize the expected confidence across all possible goals. Conditioning on the current joint belief \( j_t \), the expected reward $R(j_t)$ becomes:
\begin{align}
    R(j_t)&=\mathbb{E}_{g \sim P(g \mid j_t)}[b_t(g)] = \sum_g b_t(g) \cdot P(g \mid j_t) \\ &= \sum_g b_t^2(g)
    = (\sum_{s_t} j_t(s_t, g))^2.
\end{align}
since \( P(g \mid j_t) = b_t(g) \).
This squared belief reward encourages the observer to take actions that sharpen the goal distribution—i.e., to reduce uncertainty and increase confidence in a particular goal.

Given the reward signal defined from the joint belief, the observer aims to select actions that maximize the discounted cumulative expected reward. Formally, for any policy \(\pi^O\), we define the value function at time \(t\) as:
\begin{equation}
V_t^{\pi^O}(j_t, u_t) \;=\; \mathbb{E}^{\pi^O}\!\biggl[\sum_{k\geq t}\!\gamma^{\,k-t}\,R\bigl(j_k\bigr)\;\bigm|\;j_t,\,u_t\biggr],
\end{equation}
where \(\gamma\in[0,1]\) is the discount factor, and the expectation is taken over the stochastic belief transitions and observer‐state dynamics induced by \(\pi^O\).

Correspondingly, we define the action‐value function
\begin{equation}
Q_t^{\pi^O}(j_t, u_t, a_t) \;=\; \mathbb{E}^{\pi^O}\!\biggl[\sum_{k>t}\!\gamma^{\,k-t}\,R\bigl(j_k\bigr)\;\bigm|\;j_t,\,u_t,\,a_t\biggr],
\end{equation}
which gives the expected return when the observer takes action \(a_t\) in \((j_t,u_t)\) and thereafter follows policy \(\pi^O\).

The optimal policy \(\pi^{O*}\) is defined as the policy that selects the action maximizing the expected future value at each time step, which satisfies the Bellman relations:
\begin{align}
&V_t^{\pi^{O^*}}(j_t,u_t) 
= R(j_t) + \max_{a_t}\;Q_t^{\pi^{O^*}}(j_t,u_t,a_t), 
\\
&Q_t^{\pi^{O^*}}(j_t,u_t,a_t) \notag \\
&= \mathbb{E}[\gamma V_{t+1}^{\pi^{O^*}}(j_{t+1},u_{t+1})
|j_t,\,u_t,\,a_t]. \label{line:q_function}
\\
&\pi^{O*}(a_t\mid j_t,u_t) 
= \arg\max_{a} Q_t^{\pi^{O^*}}(j_t,u_t,a).
\end{align}

To evaluate the Q-function in Equation~\ref{line:q_function}, we decompose the expectation over the possible future observer states and observations that influence the update of the joint belief. This leads to the expansion of Equation~\ref{line:q_function}:

\begin{align}
&\mathbb{E}\left[\gamma V_{t+1}^{\pi^{O^*}}(j_{t+1}, u_{t+1}) \;\middle|\; j_t, u_t, a_t \right] \notag \\
&= \gamma \sum_{u_{t+1}} P(u_{t+1} \mid u_t, a_t)
\sum_{o_{t+1}} P(o_{t+1} \mid u_{t+1}, j_t) \notag \\
&\quad \cdot V_{t+1}^{\pi^{O^*}}(h(j_t, u_{t+1}, o_{t+1}), u_{t+1}),
\end{align}
where \( h(j_t, u_{t+1}, o_{t+1}) \) is the joint belief update function specified previously in Equation~\ref{f_transition}. The observation likelihood under the belief \( j_t \) is computed as:
\begin{align}
P(o_{t+1} \mid u_{t+1}, j_t) 
&= \sum_{s_{t+1}} P(o_{t+1} \mid s_{t+1}, u_{t+1}) P(s_{t+1}\mid j_t), \\
P(s_{t+1}\mid j_t)
&= \sum_{s_t, g} P(s_{t+1}\mid s_t, g) j_t(s_t, g).
\end{align}

This formulation highlights how the observer policy \(\pi^O\) affects the future expected reward, that is by determining the action $a_t$, the observer controls the transition to the next state $u_{t+1}$, which affects the subsequent observation $o_{t+1}$, and thereby influences the updated belief $j_{t+1}.$ Because the reward $R(j_t)$ depends on the confidence in the actor’s goal encoded in the belief, the observer is incentivized to choose actions that lead to informative observations. This captures the essence of the PAGR problem where the observer is not just passively reacting to observations but actively selecting actions to accelerate goal inference by driving belief updates toward greater certainty.

\subsection{Monte Carlo Tree Search for Active Goal Recognition}

In the previous subsection we established a principled framework for observer action selection based on maximizing the expected cumulative reward under the joint belief. This formulation, grounded in the Bellman equations, captures how observer actions influence future beliefs and ultimately the confidence in goal inference. However, computing exact value or Q-functions becomes computationally infeasible in practice due to the high-dimensional belief space and stochastic dynamics involved in belief updates. The exponential growth of possible observation-action trajectories renders exact planning intractable in realistic settings.

To overcome this challenge, we employ \textit{Monte Carlo Tree Search} (MCTS), which is a sample-based online solver that approximates optimal actions via forward simulation. Rather than exhaustively evaluating all possible belief trajectories, MCTS incrementally constructs a search tree rooted at the current belief \( j_t \) and observer state \( u_t \). Through repeated simulations, it estimates the value of different action branches by sampling possible observation outcomes and belief transitions.
At each time step, MCTS selects an action according to the approximation $\hat{a}_t \approx \arg\max_{a_t} Q_t^{\pi^{O^*}}(j_t, u_t, a_t)$, where the Q-values are estimated from simulations.

We now describe our adaptation of MCTS for the PAGR problem, which is similar to PFT-DPW introduced in \cite{sunberg2018online} but without double progressive widening.
To model the uncertainty in observations, our MCTS tree alternates between \emph{decision nodes} and \emph{chance nodes}. The root node represents the current joint belief \( j_t \) and observer state \( u_t \), and is a decision node where the observer selects an action \( a_t \). This node approximates the value function \( V_t^{\pi^{O^*}}(j_t, u_t) \), and each child node corresponds to a possible action \( a_t \), forming a chance node.
A chance node represents the Q-value \( Q_t^{\pi^{O^*}}(j_t, u_t, a_t) \). At this node, we perform the belief prediction step (Equation~\ref{line1}-\ref{line3}) and simulate forward by sampling a goal \( \hat{g} \), actor state \( s_{t+1} \), observer state \( u_{t+1} \), and observation \( o_{t+1} \) from the updated belief. Using these, we update the joint belief to obtain \( j_{t+1} = f(j_t, u_{t+1}, o_{t+1}) \) as shown in Equation~\ref{f_transition}.
We use \emph{lazy expansion} for chance nodes, meaning that new decision nodes are only generated when selected for the first time to avoid full enumeration of the large belief space.
During the tree traversal, decision nodes use the UCB1 algorithm for action selection. Chance nodes select the child by sampling from current subjective belief maintained by the observer.
In the backpropagation stage, we backup values by averaging the values of child nodes, which improves robustness under partial observability. At decision nodes, we also incorporate the immediate reward \( R(j_t) \) into the backup.
To reduce computational cost, we initialize newly expanded nodes with their immediate reward \( R(j_{t+1}) \) rather than performing random rollouts.
By combining the belief update mechanism introduced in Subsection~\ref{jbu} with MCTS described in this subsection, we obtain a complete algorithm for solving the AGR problem, which we refer to as \emph{AGR-MCTS}.

\section{Active Goal Recognition in Grid World}

We present a case study of the AGR problem in a classic two dimensional grid world domain, which is widely used in both goal recognition~\citep{masters2019cost} and active information gathering~\citep{varotto2021active}. In this environment, the actor, observer, and goals are represented as discrete positions on a grid.

While our general framework described in previous sections supports stochastic transitions and observations, we adopt a simplified deterministic setting in this section to better illustrate the core ideas. Specifically, we assume deterministic transitions for both the actor and the observer, as well as deterministic observations. As a result, the transition models (Equations~\ref{eq:act_trans} and $f^O$) and the observation function $f_{\text{obs}}$ are treated as Dirac distributions.

\begin{figure}
    \centering
    \includegraphics[width=0.5\linewidth]{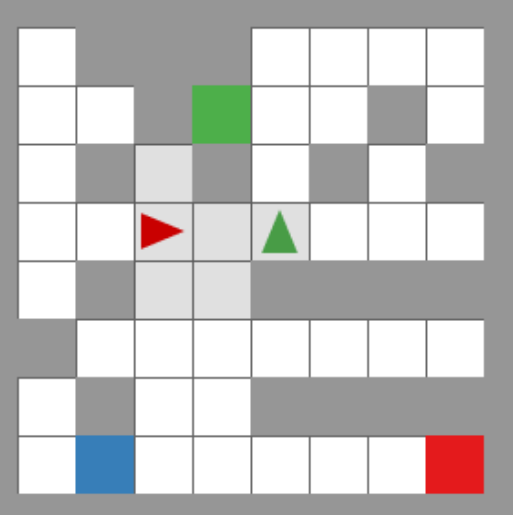}
    \caption{Illustrative experimental environment. The red cone represents the observer, and the green cone represents the actor, with each cone indicating their orientation. The light gray region denotes the observer’s field of view (FoV). Colored squares mark candidate goal locations. For clarity, the grid size is scaled down to $8\times8$ with a FoV of $3\times3$.}
    \label{fig:layout}
\end{figure}

The environment includes static obstacles that the actor cannot traverse. The actor starts from an initial location $s_0$ and aims to reach a fixed goal location $g^*$. The observer also starts from a random location $u_0$, but unlike the actor, it is allowed to traverse obstacles. Both agents share the same action space: move forward, turn left, turn right, and stay. The state of each agent is defined by its grid position and facing direction, which are the only relevant state variables in this domain. The observer also has a set of potential goals $\mathcal{G}$ so that $g^* \in \mathcal{G}$, which is consistent with prior work in goal recognition. An illustrative layout is shown in Figure~\ref{fig:layout}.

The observer is equipped with a Field of View (FoV). In our experiments, the FoV is defined as a $5\times 5$ grid extending in the observer’s facing direction. The observer is positioned at the center of the back row of this grid, meaning the FoV extends two cells to the left and right and five cells forward relative to the observer’s orientation. Cells in the FoV are excluded if a direct line of sight is blocked by an obstacle. The observation model is straightforward: the observer obtains the actor’s position if it falls within the FoV; otherwise, it receives a “not observed” signal. 

Most prior work in goal recognition assumes the actor behaves optimally, limiting applicability to realistic settings. In contrast, we introduce a more realistic and challenging scenario by modeling bounded rational behavior. Specifically, we generate a private cost map known only to the actor. The actor then computes an optimal path to its goal $g^*$ under the private cost map, resulting in behavior that may appear suboptimal from the observer’s perspective.

We implement this multi-agent environment using the MultiGrid Python library~\citep{li2023theory}.

\subsection{Experiment Setup}

We evaluate several algorithms on the previously described grid world domain to illustrate the inherent challenges of AGR. To systematically vary the difficulty of the task, we manipulate two key factors: grid size and the initial distance between the actor and the observer. 
We consider two grid sizes: a small grid ($10 \times 10$) and a large grid ($20 \times 20$). For each grid size we test  three levels of initial actor-observer distances: easy (3 cells), normal (5 cells), and hard (7 cells in small grid and 10 cells in large grid). 

The observer's Field of View (FoV) is fixed at five cells. In the easy condition, the actor often falls within the observer’s initial FoV, making it observable from the start. In contrast, under the hard condition, the actor starts far enough away that it is initially unobservable. These combinations yield a total of six unique environment configurations.

For each configuration, we generate ten random grid layouts, each with static obstacles. To simulate imperfect behavior and induce goal ambiguity, each layout includes a hidden cost map for the actor, encouraging suboptimal paths. For each layout, we then create five distinct task instances. In each instance, we randomly sample three goal candidates $\mathcal{G}$ and randomly select one as the actor’s true goal $g^*$. The actor’s starting position $s_0$ is randomly assigned. The observer’s starting position $u_0$ is then placed at the specified distance from the actor according to the configuration, and its initial orientation is set to face the actor. Importantly, the observer has access only to the goal candidate set  $\mathcal{G}$ and no other information about the actor’s internal state or behavior. Each episode concludes once the actor reaches its goal. 

\subsubsection{Evaluation Metrics.}
As introduced in Section~\ref{problem_statement}, we use \emph{convergence} (CV) as the primary evaluation metric, with a threshold parameter set to $\theta = 0.5$. In addition, we report two auxiliary metrics: (1) \emph{Final probability}: the average final belief probability assigned to the true goal $b_T(g^*)$, where $T$ denotes the length of the episode (i.e., the number of steps the actor takes to reach the goal) and (2) \emph{Success rate}: the percentage of instances in which the true goal $g^*$ was correctly identified by the observer by the end of the episode (i.e. $b_T(g^*) > \theta$ was considered as correctly identified).

\begin{table*}[ht!]
\centering
\renewcommand{\arraystretch}{1.3}
\setlength{\tabcolsep}{6pt}
\begin{tabularx}{\textwidth}{@{}p{3.2cm}X X X@{}}
\toprule
\textbf{Method} & \textbf{Goal Inference} & \textbf{Action Selection} & \textbf{Reward Design} \\
\midrule

\emph{Passive-Random} & Passive Goal Recognition & Random & N/A\\


\emph{Search-and-Follow} & Passive Goal Recognition & Actively searches for the actor, then follows its trajectory.  & N/A\\

\emph{Belief-Greedy} & Joint Belief Update & Greedy & Negative distance to most likely actor position. \\

\emph{AGR-MCTS} & Joint Belief Update & MCTS & Accumulated Goal Probability plus actor state entropy  \\





\bottomrule
\end{tabularx}
\caption{Comparison of AGR-MCTS and some of ablation algorithms, in terms of goal inference, action selection, and reward design.}
\label{tab:ablation}
\end{table*}

We compare our proposed method, \emph{AGR-MCTS}, against several baseline approaches. Here, we introduce our own domain-specific \emph{Belief-Greedy} algorithm, which uses joint belief updates but selects actions greedily. It is always moving toward the most likely actor position without considering long-term planning. A brief description of each approach is provided in Table~\ref{tab:ablation}. By default, our formulation uses the accumulated belief in the true goal as the primary reward signal, reflecting the confidence gained over time in the correct hypothesis. However, the active information gathering literature often employs entropy reduction as a reward signal to encourage uncertainty minimization~\citep{veiga_reactive_2023}. To assess the effectiveness of this alternative strategy, we extend \emph{AGR-MCTS} by incorporating an entropy-based regularization term of actor state into the reward function. We set the number of MCTS iterations to 100 to balance computational efficiency and decision quality, and the algorithm assumes an $\epsilon$-greedy actor model for belief update in Equation~\ref{eq:act_trans}.

\subsection{Passive Goal Recognition}

For the passive goal recognition algorithm, we adopt the single-observation approach proposed by \cite{masters2019cost}, which has been shown to be effective in similar grid world domains. This approach is particularly well-suited to our setting, where the number of valid observations (i.e., known actor positions) may be very limited. Its ability to perform inference based on a single observation is especially advantageous under such constraints.

The original method assumes access to the actor’s start position, which is not always available in our setting. To address this, we adapt the algorithm to a more general setting. Specifically, if no valid observation has been made, we default to the prior distribution over goals. Once a valid observation \( o \) is available, we compute an accumulated cost difference value for each goal \( g \), denoted as \( \text{cdiff}(o, g) \), and apply the probabilistic model from \cite{ramirez2010probabilistic}:
\begin{equation}
    P(g \mid o) = \alpha \cdot \frac{e^{-\beta \cdot \text{cdiff}(o, g)}}{1 + e^{-\beta \cdot \text{cdiff}(o, g)}},
\end{equation}
where \( \alpha \) is a normalization constant and \( \beta \) is a scaling parameter.

In the formulation by \cite{masters2019cost}, this cost difference is computed using the final observation and a known start state. Instead, we extend this to operate incrementally by maintaining accumulated cost differences across observations. When a new valid observation \( o_t \) is made, we consider the last valid observation \( o' \), and update $\text{cdiff}(g)$ as follows:
\begin{equation}
    \text{cdiff}(g) \leftarrow \text{cdiff}(g) + \text{optc}(o_t, g) + \text{step}(o', o_t) - \text{optc}(o', g),
\end{equation}
where \( \text{optc}(o, g) \) denotes the optimal cost from observation \( o \) to goal \( g \), which can be precomputed, and \( \text{step}(o', o_t) \) is the number of steps taken between the two observations, which is directly accessible.

This incremental formulation allows the method to accommodate suboptimal behavior and operate under partial observability, without requiring any additional online planning. Although this represents a novel adaptation of passive goal recognition for settings with substantial missing observations and online inference requirements, we present it here only briefly, as it is not the primary focus of this work.

\subsection{Results}


\begin{table*}[ht!]
\centering
\begin{tabular}{llcccccc}
\toprule
\textbf{Algorithm} & \textbf{Metric} & \textbf{S-E} & \textbf{S-N} & \textbf{S-H} & \textbf{L-E} & \textbf{L-N} & \textbf{L-H} \\
\midrule

\multirow{3}{*}{Passive-Random}
    & CV    & 0.12 & 0.03 & 0.03 & 0.21 & 0.09 & 0.06 \\
    & SR    & 0.24 & 0.12 & 0.18 & 0.28 & 0.16 & 0.12 \\
    & FP     & 0.51 & 0.42 & 0.44 & 0.52 & 0.43 & 0.40 \\
\hline

\multirow{3}{*}{Search-and-Follow}
    & CV    & 0.26 & 0.14 & 0.09 & 0.31 & 0.12 & 0.08 \\
    & SR    & 0.66 & 0.46 & 0.34 & 0.54 & 0.24 & 0.16 \\
    & FP     & 0.75 & 0.62 & 0.53 & 0.70 & 0.49 & 0.43 \\
\hline
\multirow{3}{*}{Belief-Greedy}
    & CV    & 0.35 & \textbf{0.27} & \textbf{0.24} & 0.45 & \textbf{0.31} & \textbf{0.22} \\
    & SR    & \textbf{0.82} & \textbf{0.76} & \textbf{0.68} & 0.80 & \textbf{0.70} & \textbf{0.60} \\
    & FP     & \textbf{0.87} & \textbf{0.82} & \textbf{0.78} & 0.85 & \textbf{0.79} & \textbf{0.69} \\
\hline
\multirow{3}{*}{AGR-MCTS}
    & CV    & \textbf{0.39} & 0.22 & 0.21 & \textbf{0.51} & 0.30 & 0.12 \\
    & SR    & \textbf{0.82} & 0.70 & 0.66 & \textbf{0.86} & 0.60 & 0.30 \\
    & FP     & 0.86 & 0.76 & 0.76 & \textbf{0.90} & 0.72 & 0.54 \\
\hline


\bottomrule
\end{tabular}
\caption{Results across six configurations for each algorithm. Each algorithm is evaluated using three metrics: Convergence (CV), Success Rate (SR), and Final Probability (FB). The best performance for each metric in each configuration is highlighted in \textbf{bold}. S and L denote Small and Large grid sizes, respectively, while E, N, and H indicate Easy, Normal, and Hard initial distance conditions.}
\label{tab:results_full}
\end{table*}

Table~\ref{tab:results_full} presents a comparative evaluation of the four selected algorithms across the six unique environment configurations, combining variations in grid size (Small/Large) and initial distance conditions (Easy/Normal/Hard). As the three metrics exhibit similar performance trends across configurations, we focus on the Coverage (CV) metric in the following discussion for clarity and conciseness.

Among all algorithms inspected, \emph{Belief-Greedy} consistently demonstrates strong performance across all configurations, achieving the highest scores under most conditions. \emph{AGR-MCTS}, which incorporates a more complete planning-based approach, further improves performance in easier scenarios, where it outperforms all other methods. 

In contrast, the Passive-Random and Search-and-Follow baselines perform significantly worse, indicating that non-belief-driven strategies struggle under partial observability. Overall, the results demonstrate the advantage of combining belief-aware planning and informative rewards for AGR in partially observable environments.

\subsubsection{Goal Inference Comparison} 

\begin{figure*}[ht!]
    \centering
    \includegraphics[width=0.88\linewidth]{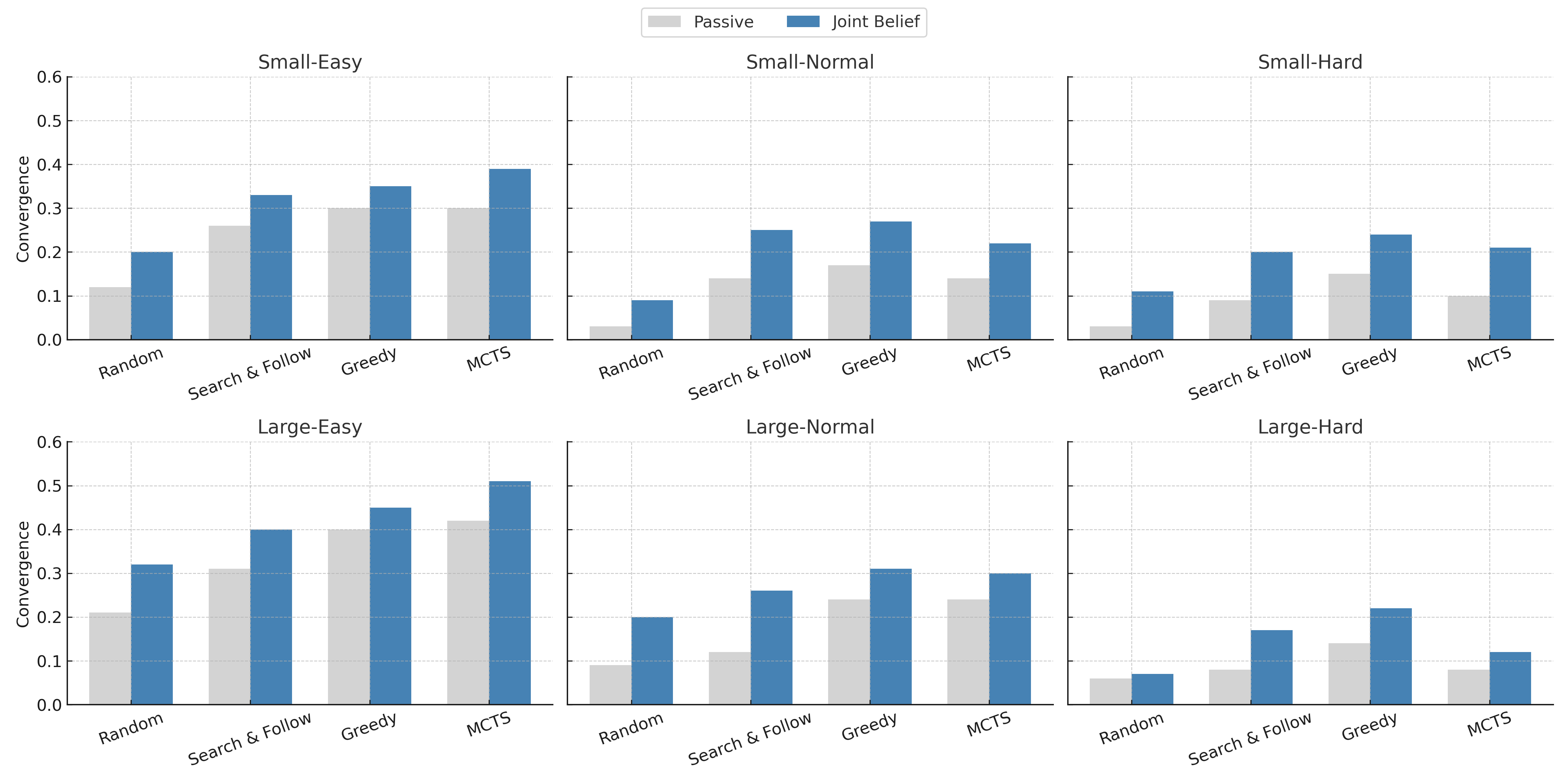}
    \caption{Comparison of CV between Joint Belief Update (active) and Passive Goal Recognition across different algorithms and scenario configurations. Each group contrasts the performance of the two inference methods under the same observation sequences. }
    \label{fig:goal-inference}
\end{figure*}

To further evaluate the effectiveness of the proposed joint belief update mechanism, we conduct an ablation study comparing it with the passive goal recognition algorithm. Specifically, for the same sequence of observations collected under four different algorithms, we apply both the passive goal recognition and the joint belief update methods, and compare their performance, as illustrated in Figure~\ref{fig:goal-inference}. 

Joint belief update outperforms the passive baseline across all configurations and action selection strategies. This improvement is primarily due to how each approach handles missing observations. In passive goal recognition, only positive observations (i.e., those where the actor is detected) are utilized for inference, while unobserved signals are treated as missing observation and thus provide no information. In contrast, the joint belief update algorithm leverages both observed and unobserved information. Returning to our motivating example, if the observer receives no signal when monitoring at the red marker, the joint belief update infers that the actor is likely pursuing an alternative goal, rather than treating this absence of evidence as uninformative.


\section{Discussion}\label{sec:discussion}

In this section, we discuss the insights from our experiment results and highlight the potential future directions in the field of AGR.

\subsection{Goal Inference Mechanism}
Our experimental evaluation confirms the superior performance of joint belief updating over passive goal recognition algorithms. Like we discussed in the previous section, the key advantage lies in the belief formulation's capacity to seamlessly integrate all available information. Although passive goal recognition methods could potentially exploit similar information by explicitly modeling missing observation effects, such an approach would incur exponential computational cost due to model size expansion. These results validate the theoretical advantages of belief-based goal inference approaches in POMDP settings.

\subsection{Action Selection Strategy}
While there is a clear advantage to the belief update and goal recognition mechanism, the complexity of designing effective online solvers in PAGR settings became apparent even within our constrained experimental domain. \emph{Belief-Greedy} enhanced with domain-specific heuristics (i.e., distance to the most probable actor position) achieved strong performance through joint belief integration, but its reliance on domain knowledge limits its broader applicability. Conversely, our MCTS-based approach, though not consistently superior to the greedy algorithm, maintains complete domain independence, thereby offering greater generalizability across different domains. 

For completeness, we also tested MCTS using the same heuristic-based reward signal, yet it still underperformed compared to the theoretically myopic greedy algorithm. One possible explanation is that MCTS fails to search deep enough to gather meaningful reward signals. In our domains, rewards are extremely sparse, because a strong signal only arises when the observer detects the actor. Indeed, we measured the average search depths of just four steps in large grids (and eight in small ones), explaining why MCTS only outperforms the greedy strategy in simpler scenarios where positive observations are more likely. These findings suggest enhancing online solvers to better exploit belief information for tree pruning and deeper lookahead (e.g., techniques like double progressive widening~\citep{sunberg2018online}) as an important avenue for future work.


\subsection{Relaxing Keyhole Assumption}
As mentioned previously, this work focuses exclusively on keyhole goal recognition~\citep{ijcai2021p615}, where the actor remains completely unaffected by the observer. While this assumption holds in certain scenarios, such as a high-altitude drone monitoring ground targets, it limits the applicability to other real-world situations. An important future direction is extending this framework to cases where the actor is aware of the observer, whether in collaborative settings (e.g., transparent planning as formulated in \cite{macnally2018action}) or adversarial contexts. This represents a crucial step toward more interactive multi-agent environments.

\section{Conclusion}

In this paper, we introduced Probabilistic Active Goal Recognition (PAGR), a novel probabilistic formulation of the Active Goal Recognition problem. Our main contributions are twofold: (1) a formal definition of PAGR grounded in the Partially Observable Markov Decision Process (POMDP) framework, which enables principled reasoning under uncertainty; and (2) an integrated solution framework that combines joint belief update with Monte Carlo Tree Search (MCTS) to efficiently solve the problem without relying on domain-specific knowledge.

We developed a comprehensive set of baselines to empirically evaluate the effectiveness of our approach. The joint belief update was shown to significantly outperform passive goal recognition methods by making more effective use of available information. Additionally, our domain-independent MCTS approach performed comparable to that of our strong domain-specific greedy algorithm, suggesting a promising direction for future work on developing more effective domain-independent methods for PAGR.

In summary, the presented contributions push the boundaries of AGR and offer a rigorous platform for continued exploration in multi-agent reasoning, goal inference, and decision-making under uncertainty.

\section*{Acknowledgments}
This work is supported by the DARPA Assured Neuro Symbolic Learning and Reasoning (ANSR) program under award number FA8750-23-2-1016.

\setlength{\bibhang}{0pt}
\bibliographystyle{kr}
\bibliography{kr-sample}

\begin{thebibliography}{}

\bibitem[\protect\citeauthoryear{Amato and Baisero}{2019}]{amato2019active}
Amato, C., and Baisero, A.
\newblock 2019.
\newblock Active goal recognition.
\newblock {\em arXiv preprint arXiv:1909.11173}.

\bibitem[\protect\citeauthoryear{Araya \bgroup et al\mbox.\egroup }{2010}]{NIPS2010_68053af2}
Araya, M.; Buffet, O.; Thomas, V.; and Charpillet, F.
\newblock 2010.
\newblock A pomdp extension with belief-dependent rewards.
\newblock In Lafferty, J.; Williams, C.; Shawe-Taylor, J.; Zemel, R.; and Culotta, A., eds., {\em Advances in Neural Information Processing Systems}, volume~23.
\newblock Curran Associates, Inc.

\bibitem[\protect\citeauthoryear{Bajcsy}{1988}]{bajcsy_active_1988}
Bajcsy, R.
\newblock 1988.
\newblock Active perception.
\newblock {\em Proceedings of the IEEE} 76(8):966--1005.

\bibitem[\protect\citeauthoryear{Canese \bgroup et al\mbox.\egroup }{2021}]{canese2021multi}
Canese, L.; Cardarilli, G.~C.; Di~Nunzio, L.; Fazzolari, R.; Giardino, D.; Re, M.; and Span{\`o}, S.
\newblock 2021.
\newblock Multi-agent reinforcement learning: A review of challenges and applications.
\newblock {\em Applied Sciences} 11(11):4948.

\bibitem[\protect\citeauthoryear{Dann \bgroup et al\mbox.\egroup }{2023}]{dann2023multi}
Dann, M.; Yao, Y.; Alechina, N.; Logan, B.; Meneguzzi, F.; and Thangarajah, J.
\newblock 2023.
\newblock Multi-agent intention recognition and progression.
\newblock In {\em Proceedings of the 32nd International Joint Conference on Artificial Intelligence, IJCAI 2023},  91--99.
\newblock IJCAI Organization.

\bibitem[\protect\citeauthoryear{Demiris}{2007}]{demiris2007prediction}
Demiris, Y.
\newblock 2007.
\newblock Prediction of intent in robotics and multi-agent systems.
\newblock {\em Cognitive processing} 8(3):151--158.

\bibitem[\protect\citeauthoryear{Fitzpatrick \bgroup et al\mbox.\egroup }{2021}]{fitzpatrick2021behaviour}
Fitzpatrick, G.; Lipovetzky, N.; Papasimeon, M.; Ramirez, M.; and Vered, M.
\newblock 2021.
\newblock Behaviour recognition with kinodynamic planning over continuous domains.
\newblock {\em Frontiers in Artificial Intelligence} 4:717003.

\bibitem[\protect\citeauthoryear{Gall, Ruml, and Keren}{2021}]{ijcai2021p559}
Gall, K.~C.; Ruml, W.; and Keren, S.
\newblock 2021.
\newblock Active goal recognition design.
\newblock In Zhou, Z.-H., ed., {\em Proceedings of the Thirtieth International Joint Conference on Artificial Intelligence, {IJCAI-21}},  4062--4068.
\newblock International Joint Conferences on Artificial Intelligence Organization.
\newblock Main Track.

\bibitem[\protect\citeauthoryear{Kaminka, Vered, and Agmon}{2018}]{kaminka2018plan}
Kaminka, G.; Vered, M.; and Agmon, N.
\newblock 2018.
\newblock Plan recognition in continuous domains.
\newblock In {\em Proceedings of the AAAI Conference on Artificial Intelligence}, volume~32.

\bibitem[\protect\citeauthoryear{Li \bgroup et al\mbox.\egroup }{2023}]{li2023theory}
Li, H.; Chong, Y.; Stepputtis, S.; Campbell, J.~P.; Hughes, D.; Lewis, C.; and Sycara, K.
\newblock 2023.
\newblock Theory of mind for multi-agent collaboration via large language models.
\newblock In {\em Proceedings of the 2023 Conference on Empirical Methods in Natural Language Processing},  180--192.

\bibitem[\protect\citeauthoryear{MacNally \bgroup et al\mbox.\egroup }{2018}]{macnally2018action}
MacNally, A.~M.; Lipovetzky, N.; Ramirez, M.; and Pearce, A.~R.
\newblock 2018.
\newblock Action selection for transparent planning.
\newblock In {\em Proceedings of the 17th International Conference on Autonomous Agents and MultiAgent Systems},  1327--1335.

\bibitem[\protect\citeauthoryear{Masters and Sardina}{2019}]{masters2019cost}
Masters, P., and Sardina, S.
\newblock 2019.
\newblock Cost-based goal recognition in navigational domains.
\newblock {\em Journal of Artificial Intelligence Research} 64:197--242.

\bibitem[\protect\citeauthoryear{Masters and Vered}{2021}]{ijcai2021p615}
Masters, P., and Vered, M.
\newblock 2021.
\newblock What’s the context? implicit and explicit assumptions in model-based goal recognition.
\newblock In Zhou, Z.-H., ed., {\em Proceedings of the Thirtieth International Joint Conference on Artificial Intelligence, {IJCAI-21}},  4516--4523.
\newblock International Joint Conferences on Artificial Intelligence Organization.
\newblock Survey Track.

\bibitem[\protect\citeauthoryear{Meneguzzi and Fraga~Pereira}{2021}]{meneguzzi_survey_2021}
Meneguzzi, F., and Fraga~Pereira, R.
\newblock 2021.
\newblock A {Survey} on {Goal} {Recognition} as {Planning}.
\newblock In {\em Proceedings of the {Thirtieth} {International} {Joint} {Conference} on {Artificial} {Intelligence}},  4524--4532.
\newblock Montreal, Canada: International Joint Conferences on Artificial Intelligence Organization.

\bibitem[\protect\citeauthoryear{Oliehoek, Amato, and others}{2016}]{oliehoek2016concise}
Oliehoek, F.~A.; Amato, C.; et~al.
\newblock 2016.
\newblock {\em A concise introduction to decentralized POMDPs}, volume~1.
\newblock Springer.

\bibitem[\protect\citeauthoryear{Ram{\'\i}rez and Geffner}{2010}]{ramirez2010probabilistic}
Ram{\'\i}rez, M., and Geffner, H.
\newblock 2010.
\newblock Probabilistic plan recognition using off-the-shelf classical planners.
\newblock In {\em Proceedings of the AAAI conference on artificial intelligence}, volume~24,  1121--1126.

\bibitem[\protect\citeauthoryear{Shah}{2014}]{shah2014collaborative}
Shah, C.
\newblock 2014.
\newblock Collaborative information seeking.
\newblock {\em Journal of the Association for Information Science and Technology} 65(2):215--236.

\bibitem[\protect\citeauthoryear{Shi \bgroup et al\mbox.\egroup }{2025}]{shi2025muma}
Shi, H.; Ye, S.; Fang, X.; Jin, C.; Isik, L.; Kuo, Y.-L.; and Shu, T.
\newblock 2025.
\newblock Muma-tom: Multi-modal multi-agent theory of mind.
\newblock In {\em Proceedings of the AAAI Conference on Artificial Intelligence}, volume~39,  1510--1519.

\bibitem[\protect\citeauthoryear{Shvo and McIlraith}{2020}]{shvo2020active}
Shvo, M., and McIlraith, S.~A.
\newblock 2020.
\newblock Active goal recognition.
\newblock In {\em Proceedings of the AAAI Conference on Artificial Intelligence}, volume~34,  9957--9966.

\bibitem[\protect\citeauthoryear{Silver and Veness}{2010}]{silver2010monte}
Silver, D., and Veness, J.
\newblock 2010.
\newblock Monte-carlo planning in large pomdps.
\newblock {\em Advances in neural information processing systems} 23.

\bibitem[\protect\citeauthoryear{Stuart and Norvig}{2016}]{stuart2016artificial}
Stuart, R., and Norvig, P.
\newblock 2016.
\newblock Artificial intelligence: a modern approach (global edition).
\newblock {\em Harlow: Pearson}.

\bibitem[\protect\citeauthoryear{Sunberg and Kochenderfer}{2018}]{sunberg2018online}
Sunberg, Z., and Kochenderfer, M.
\newblock 2018.
\newblock Online algorithms for pomdps with continuous state, action, and observation spaces.
\newblock In {\em Proceedings of the International Conference on Automated Planning and Scheduling}, volume~28,  259--263.

\bibitem[\protect\citeauthoryear{{Thomas Vincent}, {Hutin Gérémy}, and {Buffet Olivier}}{2020}]{thomas_vincent_monte_2020}
{Thomas Vincent}; {Hutin Gérémy}; and {Buffet Olivier}.
\newblock 2020.
\newblock Monte {Carlo} {Information}-{Oriented} {Planning}.
\newblock In {\em Frontiers in {Artificial} {Intelligence} and {Applications}}. IOS Press.

\bibitem[\protect\citeauthoryear{Van-Horenbeke and Peer}{2021}]{van-horenbeke_activity_2021}
Van-Horenbeke, F.~A., and Peer, A.
\newblock 2021.
\newblock Activity, {Plan}, and {Goal} {Recognition}: {A} {Review}.
\newblock {\em Frontiers in Robotics and AI} 8:643010.

\bibitem[\protect\citeauthoryear{Varotto, Cenedese, and Cavallaro}{2021}]{varotto2021active}
Varotto, L.; Cenedese, A.; and Cavallaro, A.
\newblock 2021.
\newblock Active sensing for search and tracking: A review.
\newblock {\em arXiv preprint arXiv:2112.02381}.

\bibitem[\protect\citeauthoryear{Veiga and Renoux}{2023}]{veiga_reactive_2023}
Veiga, T., and Renoux, J.
\newblock 2023.
\newblock From {Reactive} to {Active} {Sensing}: {A} {Survey} on {Information} {Gathering} in {Decision}-theoretic {Planning}.
\newblock {\em ACM Computing Surveys} 55(13s):1--22.

\bibitem[\protect\citeauthoryear{Vered and Kaminka}{2017}]{vered2017heuristic}
Vered, M., and Kaminka, G.~A.
\newblock 2017.
\newblock Heuristic online goal recognition in continuous domains.
\newblock {\em arXiv preprint arXiv:1709.09839}.

\bibitem[\protect\citeauthoryear{Vered \bgroup et al\mbox.\egroup }{2018}]{vered2018towards}
Vered, M.; Pereira, R.~F.; Kaminka, G.; and Meneguzzi, F.~R.
\newblock 2018.
\newblock Towards online goal recognition combining goal mirroring and landmarks.
\newblock In {\em Proceedings of the 19th International Conference on Autonomous Agents and Multiagent Systems, 2018, Su{\'e}cia.}

\bibitem[\protect\citeauthoryear{Vered, Kaminka, and Biham}{2016}]{vered2016online}
Vered, M.; Kaminka, G.~A.; and Biham, S.
\newblock 2016.
\newblock Online goal recognition through mirroring: Humans and agents.
\newblock In {\em Annual Conference on Advances in Cognitive Systems 2016}.
\newblock Cognitive Systems Foundation.

\bibitem[\protect\citeauthoryear{Zhang, Kemp, and Lipovetzky}{2023}]{zhang2023goal}
Zhang, C.; Kemp, C.; and Lipovetzky, N.
\newblock 2023.
\newblock Goal recognition with timing information.
\newblock In {\em Proceedings of the international conference on automated planning and scheduling}, volume~33,  443--451.

\bibitem[\protect\citeauthoryear{Zhang, Yang, and Ba{\c{s}}ar}{2021}]{zhang2021multi}
Zhang, K.; Yang, Z.; and Ba{\c{s}}ar, T.
\newblock 2021.
\newblock Multi-agent reinforcement learning: A selective overview of theories and algorithms.
\newblock {\em Handbook of reinforcement learning and control}  321--384.

\end{thebibliography}

\end{document}